\documentclass[10pt,twocolumn,letterpaper]{article}

\usepackage{iccv}
\usepackage{times}
\usepackage{epsfig}
\usepackage{graphicx}
\usepackage{amsmath}
\usepackage{amssymb}
\usepackage{multirow}

\usepackage[pagebackref=true,breaklinks=true,letterpaper=true,colorlinks,bookmarks=false]{hyperref}

\iccvfinalcopy 

\def\iccvPaperID{5666} 

\ificcvfinal\fi
\begin{document}

\title{DASNet: Reducing Pixel-level Annotations for Instance and Semantic Segmentation}

\author{Chuang Niu\\
Xidian University\\
{\tt\small niuchuang@stu.xidian.edu.cn}
\and
Shenghan Ren\\
Xidian University\\
{\tt\small shren@xidian.edu.cn}
\and
Jimin Liang\\
Xidian University\\
{\tt\small jimleung@mail.xidian.edu.cn}
}

\maketitle

\begin{abstract}
Pixel-level annotation demands expensive human efforts and limits the performance of deep networks that usually benefits from large-scale training data. In this work we aim to achieve high quality instance and semantic segmentation results using only a small set of accurate pixel-level mask annotations and a large set of rough box annotations. The basic idea is simplifying the pixel-level supervised learning task by exploring the detection models, and reducing the required amount of mask annotations. Our architecture, named DASNet, consists of three modules: detection, attention, and segmentation. The detection module detects all classes of objects, the attention module generates multi-scale class-specific features, and the segmentation module recovers the binary masks. Our method demonstrates substantially improved performance compared to existing semi-supervised approaches on PASCAL VOC 2012 dataset.
\end{abstract}

\section{Introduction}

Deep convolutional neural networks (DCNN) have been proven to be an effective solution to pattern recognition problems in computer vision. Generally, DCNN usually requires extensive amounts of data to train a robust and accurate model. Particularly, modern datasets \cite{Deng2009ImageNet}, \cite{Krishna2017Visual}, \cite{openimages} consist of millions of annotated images that cover thousands of object categories in image tags or bounding boxes (bbox), and dramatically improved the classification and detection performance. In contrast, the common datasets \cite{Everingham2010The}, \cite{Lin2014Microsoft} have orders of magnitudes fewer pixel-level mask annotations, since acquiring mask annotations requires much more human efforts (about 15$\times$ more time than box annotations \cite{Lin2014Microsoft}). This limits the performance of instance and semantic segmentation models that usually benefit from the large-scale pixel-level annotated training data.

In the past few years, weakly- or semi-supervised learning methods are explored to mitigate the problem by utilizing available large-scale weak annotations, such as image-level labels \cite{Pathak2014}, \cite{Pedro2015}, \cite{Pathak2016}, \cite{Saleh2016}, \cite{Papandreou2015}, \cite{Lu2017}, \cite{Wei2015a}, \cite{Wei2015b}, \cite{Kwak2017}, \cite{Chaudhry2017Discovering}, \cite{Roy2017Combining}, \cite{Durand2017WILDCAT}, \cite{Oh2017Exploiting} and bboxes \cite{Papandreou2015}, \cite{Dai2015}, \cite{Khoreva2017}. They often convert the weak annotations to pixel-level supervision with unsupervised approaches and train the fully convolutional networks (FCNs) \cite{Long2015Fully}, \cite{Chen2014Semantic}  by iteratively inferring and refining segmentation mask. However, this is actually a label denoising or data augmentation process and does not reduce the requirements of pixel-level training data by the segmentation models. In addition, the weak information cannot be effectively and accurately utilized because the converted mask labels are usually noisy or incomplete. Therefore, the large number of available weak annotations cannot be fully used for segmentation task.

In this work, we aim to achieve high quality instance and semantic segmentation results using only a small set of accurate pixel-level mask annotations and a large set of rough box annotations, as shown in Fig. \ref{fig:seg} and Fig. \ref{fig:inst}.  Our architecture, named DASNet, consists of three modules for detection, attention and segmentation, respectively. The detection module is to recognize and localize all objects of each class in bboxes. Given the products of detection, the attention module aims to generate multi-scale class-specific features which are used as the input to the segmentation module. The segmentation module trained with the pixel-level annotations outputs binary segmentation masks. In order to achieve instance segmentation, the position-sensitive score map technique \cite{Li2016Fully} is carefully adapted to the proposed DASNet.

Technically, we have two contributions. First, different from conventional ROI (region-of-interest) pooling strategy for object detection, our attention mechanism generates the same size of class-specific feature map by zeroing out unrelated signals only. Therefore, the whole spatial information of all class-specific object instances can be preserved. Second, compared with the DecoupleNet that takes the class-specific features from the most top layer only, our designed segmentation model gradually merges multi-scale class-specific features from multiple layers and is more robust to scale variation. By carefully adapting the position-sensitive scoring operation to our designed binary segmentation module, instance segmentation can be achieved.

Our framework has two major advantages. First, the class-agnostic segmentation strategy dramatically simplifies the pixel-level supervised learning task, which is to recover object shapes only. And all classes share the same segmentation model. Therefore, the segmentation model can easily converge to good local minima even trained with a small number of pixel-level annotations. Second, the state-of-the-art object detection models are deployed to achieve high quality object recognition and coarse localization for segmentation. Instead converted to the latent pixel-level labels, available large scale bboxes are utilized in an effective and accurate manner by training the detection model. In this context, the key problem is how to leverage the high-level detection products to further facilitate the low-level mask recovering. We solve this problem by an attention mechanism. The experimental results on PASCAL VOC 2012 dataset show that our architecture substantially outperforms existing weakly- and semi-supervised techniques especially using a small number of mask annotations.

\section{Related Work}
Since acquiring pixel-level mask is an expensive, time-consuming annotation work, researchers have recently pay more attention to develop techniques for achieving high quality segmentation results when training examples with mask annotations are limited or missing. These techniques can be roughly classified as weakly- and semi-supervised learning. We discuss each in turn next.

\textbf{Weakly-supervised learning:} In order to avoid the constraints of expensive pixel-level annotations, weakly-supervised approaches train the segmentation models with only weak annotations, including image-level labels \cite{Pathak2014}, \cite{Pedro2015}, \cite{Pathak2016}, \cite{Saleh2016}, \cite{Papandreou2015}, \cite{Lu2017}, \cite{Wei2015a}, \cite{Wei2015b}, \cite{Kwak2017}, \cite{Chaudhry2017Discovering}, \cite{Roy2017Combining}, \cite{Durand2017WILDCAT}, \cite{Oh2017Exploiting}, points \cite{Bearman2016}, scribbles \cite{Xu2015}, \cite{Lin2016}, and  bboxes \cite{Papandreou2015}, \cite{Dai2015}, \cite{Khoreva2017}. All of these methods are dedicated to converting the weak annotations to the latent pixel-level supervision and training the FCNs by iteratively inferring and refining segmentation mask. However, the latent supervisions are either incomplete or noisy without guiding with any strong annotations. Therefore, the segmentation performance of these techniques is still inferior.

\textbf{Semi-supervised leaning:} In semi-supervised learning of semantic segmentation, the assumption is that a large number of weak annotations and a small number of strong (pixel-level) annotations are available, which is usually satisfied in practice. Various types of weak annotations, such as image-level labels \cite{Hong2015Decoupled}, \cite{Papandreou2015}, scribbles \cite{Lin2016}, and bboxes \cite{Dai2015}, \cite{Papandreou2015}, have been explored in the semi-supervised setting. Intuitively, \cite{Papandreou2015}, \cite{Lin2016}, \cite{Dai2015} augment the weak annotations with the small number of strong annotations for training the FCNs as in the weakly-supervised settings and achieve better performance than the weakly supervised counterparts. In contrast, \cite{Hong2015Decoupled} decouples the semantic segmentation into two sub-tasks: classification and class-agnostic segmentation, supervised with image-level labels and pixel-level masks respectively. This approach shows an impressive performance even with a very small number of strong annotations.

It is noted that the common pre-training strategy (e.g., pre-training on ImageNet \cite{Deng2009ImageNet} for classification task) in fully supervised setting can be also regarded as a way for alleviating the requirements of strong annotations. However, its main role is to promote the convergence of the segmentation models \cite{Long2015Fully}.

In this work, we assume that there exists a large number of box annotations and a small number of mask annotations. Our method is most related to \cite{Hong2015Decoupled}, \cite{Dai2015}, \cite{Khoreva2017}. The box annotations in \cite{Dai2015}, \cite{Khoreva2017} are converted to mask labels using unsupervised methods \cite{Uijlings2013Selective}, \cite{Pont2015}, \cite{Rother2004} and a priori knowledge for training FCNs. In contrast, we utilize box annotations in an efficient and accurate manner by training the detection module. Motivated by \cite{Hong2015Decoupled}, we also decouple the semantic segmentation into sub-tasks. However, the class-specific features generated from the classification component in \cite{Hong2015Decoupled} are usually sparse and noisy, since the classification network tends to focus only on small discriminative parts (e.g. the head of an animal). In this work, by exploring the detection model, multiple complete objects of multiple scales are focused on the class-specific features with an attention module.

\section{Methods}

\subsection{Architecture Overview}

\begin{figure}
\centering
\includegraphics[height=7cm]{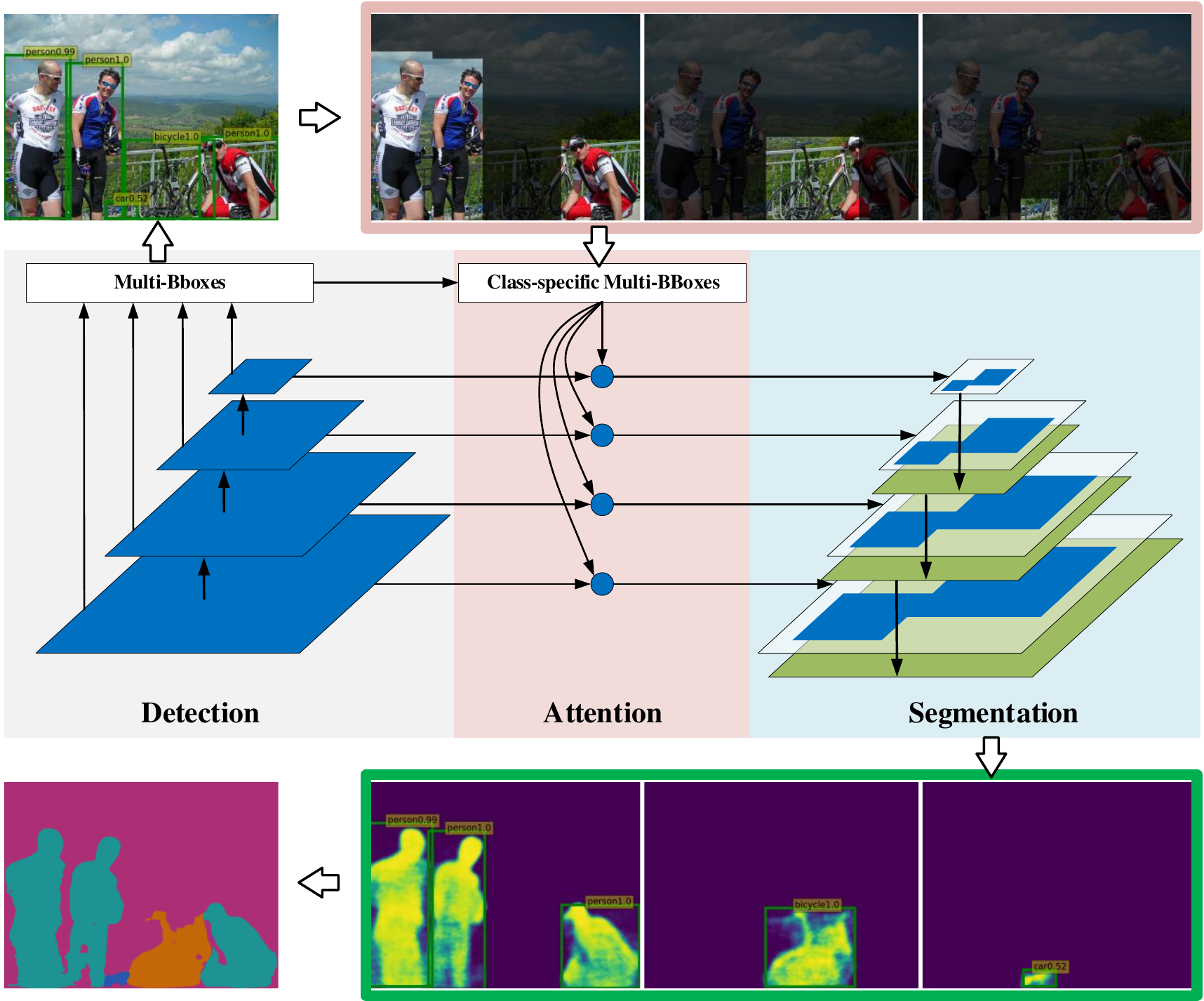}
\caption{Architecture Overview.}
\label{fig:arc}
\end{figure}

Our proposed network consists of three modules: detection, attention, and segmentation, as shown in Fig. \ref{fig:arc}. Given an image, the detection model first detects all classes of objects with bboxes. Then the attention module generates class-specific multi-scale features by cropping the multiple layer features of the detection module with the detected boxes. Finally, the segmentation module integrates all scales of feature maps hierarchically and recovers the binary masks for each class separately. Moreover, we demonstrate that the DASNet can achieve instance segmentation by computing position-sensitive score maps \cite{Li2016Fully}. Next, we will elaborate each module as well as the training and inference processes.

\subsection{Detection Module}

Training with large amounts of bbox annotations, object detection based on DCNN have achieved significant advances in recent years. In this paper, we employ the SSD models \cite{Liu2016SSD} for our object detection purpose. One of main properties for SSD is that multiple feature maps with different resolutions are combined to handle various sizes of objects, which has become a common technique. It is worth noting that we do not see any hurdles to prevent other state-of-the-art detection models from being integrated into DASNet. To facilitate understanding, we briefly review the SSD approach below.

The SSD architecture is a single feed-forward convolutional network, which consists of a base network and multiple small convolutional predictors. Given an image, the base network generates multi-scale feature maps, noted as $F=\{f_k\}$, $k\in [1, m]$, where $m$ is the number of different scale feature maps. On each feature map, a convolutional predictor is applied to produce a fixed-size collection of boxes and the corresponding object class scores of instances presented in those boxes. All scored boxes are further processed by a non-maximum suppression algorithm to produce the final detections, noted as $B=\{(x_{min}^b, y_{min}^b, x_{max}^b, y_{max}^b, l^b)\}$, where $x_{min}^b, y_{min}^b, x_{max}^b, y_{max}^b$ and $l$ are coordinates and class label of the detected box $b$. Refer to \cite{Liu2016SSD} for more details.

Therefore the detection module in the DASNet produces two kinds of products: multi-scale feature maps $F$ and detection results $B$. In the following, we will introduce how these products are utilized for reducing the pixel-level training data requirements for instance and semantic segmentation.

\subsection{Attention Module}

Correspondence property of convolutional features \cite{Long2014Do} is one of the main reasons that DCNN can be successfully applied for localization tasks. Specifically, the convolutional features of a particular layer can be regarded as a two-dimensional grid of feature vectors or a feature map. The correspondence refers that if feature vectors in a particular feature map have similar values (e.g., in inner product), their corresponding receptive field regions of the original image will have similar appearances and vice versa. Thus the vector values $f(x, y)$ and their position information (i.e., $(x, y)$) represent "what" and "where", respectively.

Motivated by this property, we design a simple yet effective attention mechanism to generate class-specific muti-scale features by exploring the products of detection module. Formally, given the target class $l$ and the detection products $F$ and $B$, a set of class-specific boxes $B^l=\{(x_{min}^b, y_{min}^b, x_{max}^b, y_{max}^b, l)\}$ are selected at first. Then the class-specific multi-scale features $F^l=\{f_k^l\}$ are computed by cropping each scale of feature map with the class-specific boxes as equation (\ref{eq_att}):
\begin{equation}
\label{eq_att}
f_k^l(x,y)=\left\{
             \begin{array}{lc}
             f_k(x,y), & x \in [W_k x_{min}^b, W_k x_{max}^b],\\
             & y \in [H_k y_{min}^b, H_k y_{max}^b], \\
             0, & otherwise
             \end{array}
\right.
\end{equation}
where $x$, $y$, and $H_k$, $W_k$ represent the feature vector coordinates and the size of feature map,respectively. The box coordinates $x_{min}^b$, $y_{min}^b$, $x_{max}^b$, $y_{max}^b$ are normalized by the image size. Briefly, the class-specific feature maps are obtained by setting the feature vector values of detection module outside all class-specific boxes as zeros. The backward process of the attention operation has similar form.

Significant differences between ROI (region-of-interest) pooling \cite{Girshick2015Fast} and our attention mechanism should be noted, although they share similar forms.
The ROI pooling takes an a feature map and a bbox as the input and outputs a small fixed-size dimension of feature map by pooling the inside feature vectors of the single box region. Since all other outside features are removed, the whole spatial information in ROI pooling is lost. On the contrary, our attention module generates the same size of class-specific feature map by zeroing out unrelated signals only. Therefore, the whole spatial information of all class-specific object instances can be preserved. This information is necessary for recovering the high-resolution object masks with the deconvolutional segmentation network discussed next.

Three properties of the multi-scale class-specific feature maps facilitate the pixel-level supervised learning task. (1) A particular object class of the feature maps has been identified in advance. (2) The object instances has been coarsely localized in high-level space by suppressing unrelated signals. (3) Multi-layer feature maps contain rich information for capturing various size of objects by leveraging the detection results.

\subsection{Segmentation Module}

Given a set of muti-scale class-specific feature maps, $F^l=\{f_k^l\}$, generated by the attention module, the objective of segmentation module is to produce binary shape masks of class-specific object instances. In semantic segmentation task, it is a pixel-wise binary classification problem which infers whether each pixel belongs to the given class $l$ or not. Furthermore, the instance segmentation needs to determine not only whether each pixel belongs to the given class, but also which particular instance it belongs to, as shown in Fig. \ref{fig:inst}.

\subsubsection{Semantic Segmentation}
The segmentation module is a single deconvolution network, which has been successfully applied for the semantic segmentation task \cite{Noh2015Learning}, \cite{Badrinarayanan2017SegNet}, \cite{2018arXiv180202611C}. Similar to \cite{Hong2015Decoupled}, we employ the deconvolution network proposed in \cite{Noh2015Learning} as our segmentation module. As shown in Fig. \ref{fig:arc}, the segmentation module hierarchically merges multi-scale class-specific feature maps in a top-down manner and outputs a segmentation mask in the same size to the input image. Specifically, to merge class-specific feature maps $f_m$ and $f_{m-1}$ ($f_m$ has the smallest size), $f_m$ is fed into a series of deconvolution and/or unpooling layers to generate a upsampled feature map $f_{m-1}^{up}$ in the same size to $f_{m-1}$ (including height, width, and channel number). Then $f_{m-1}^{up}$ and $f_{m-1}$ are concatenated along their channel direction. In turn, the concatenated feature map $f_{m-1}^{concat}$ will be merged with the lower feature map $f_{m-2}$ in the same way. This process is repeated until all class-specific feature maps are merged and the segmentation mask is generated.

For semantic segmentation task, the segmentation module produces a two-channel class-specific segmentation map, in which the two channels represent foreground and background respectively. The class-specific segmentation loss is the softmax loss over two binary class (foreground and background) in pixel-wise.

Benefit from the properties (1) and (2) of class-specific feature maps aforementioned, the pixel-level supervised learning task of detection module has been dramatically simplified to determine whether each pixel within the region of class-specific boxes belongs to a given class. Therefore, the segmentation accuracy is still competitive even training with a very small number of pixel-level annotated training samples (e.g., 5 to 10 annotations per class) as demonstrated in section \ref{sec:semantic}. Moreover, the property (3) allows the segmentation module to capture objects of various scales, see Fig. \ref{fig:seg}.

\subsubsection{Instance Segmentation}

In this section, we show that how the position-sensitive score map technique \cite{Li2016Fully} is adapted to DASNet for achieving instance segmentation.
We begin by introducing the original position-sensitive score map approach. From the top convolution features ($conv5$), $2k^2 \times (C+1)$ position-sensitive score maps are produced, where C is object class number (+1 for background), k is the size of relative position grid, and 2 presents two groups (inside and outside). Given a ROI generated by the region proposal network \cite{Ren2017Faster}, its pixel-wise score maps are produced by the assembling (copy-paste) its $k \times k$ cells from the corresponding score maps. For each pixel in a ROI, there are two tasks: 1) detection: whether it belongs to an object bbox at a relative position or not; 2) segmentation: whether it is inside an object instance's boundary or not. The detection score of the whole ROI is obtained via average pooling over all pixels' likelihoods, which are the max values between their inside and outside scores. The segmentation score (in probabilities) is the union of pixel-wise inside/outside softmax values. It is noted that the detection and segmentation scores are computed for each category. Thus, for each ROI, a softmax detection loss over $C+1$ categories, a softmax segmentation loss within the class-spcific bboxes area in foreground mask of the ground-truth category only, and a bbox regression \cite{Girshick2015Fast} loss are applied for training.

To integrate position-sensitive score map approach in DASNet, some important modifications are necessary. \textbf{\emph{First}}, the number of position-sensitive score map is reduced to $2k^2$ ($k=7$ by default in the following experiments). Since the binary class-specific segmentation of DASNet assumes that object instances of a particular class have been detected by detection module. Therefore, the segmentation module only needs to segment inside/outside masks within the detected bbox. \textbf{\emph{Second}}, the position-sensitive score maps are produced from the top convolutional features ($deconv1$) with high-resolution of a deconvolotion network, instead of $conv5$ feature map with much ($16\times$) lower resolution of a truncated network. Its effectiveness has not been proven in this case before. \textbf{\emph{Third}}, in the training stage, each bbox has two loss terms in equal weights: an instance score regression sigmoid-cross-entropy loss and a softmax segmentation loss over the foreground mask of the ground-truth instance only. \textbf{\emph{Fourth}}, in the testing stage, the segmentation module only outputs the detected instance mask. It is noted that the instance score regression loss can not be removed, otherwise the learned score maps are not position-sensitive without learning negative instance samples. Thus, it should be careful to organize the training samples for learning position-sensitive score maps, which will be discussed next.

\subsection{Training and Inference}

\textbf{Stage-wise training vs. joint training:}
For stage-wise training, we first train the detection module with bbox annotations, and then train the segmentation module with mask annotations by freezing the parameters of the detection module. For joint training, since the attention module allows the gradients from segmentation module backward to the detection module, the detection module and segmentation module are trained with the bboxes and mask annotations simultaneously. Fine-tuning strategy is another option that simultaneously fine-tuning the models obtained from stage-wise training. However, we do not see improvements with either joint training or fine-tuning strategy by now. Therefore, all experiments are conducted with the stage-wise training in this paper.

\textbf{Detection:}
The training and inference processes of detection module are same as \cite{Liu2016SSD}.

\textbf{Semantic segmentation:}
Both mask and bbox annotations (bbox annotations can be easily obtained from mask annotations) are needed to train the segmentation module. At the training stage, the ground-truth class-specific bboxes are fed to the attention module for generating multi-scale class-specific features, which are the input to the segmentation module. And the ground-truth binary masks corresponding to the class-specific bboxes are used to compute the segmentation loss, in which the pixels outside class-specific bboxes are ignored. In inference, the final semantic segmentation mask is obtained via a max operation on all class-specific score maps.

\textbf{Instance segmentation:}
To train the segmentation module for instance segmentation, instance-aware semantic segmentation mask and bbox annotations are required. In this setting, the bbox annotations are utilized in two ways: 1) The ground-truth class-specific bboxes are used to generate multi-scale class-specific features. 2) For each ground-truth bbox, $p$ positive and $n$ negative instance bboxes ($p=2, n=4$ in our experiments) are sampled for training the position-sensitive score maps. Specifically, the sampled bboxes which match one of ground-truth bboxes (Jaccard overlap larger than 0.5) are treated as positives, and those do not match any of ground-truth bboxes are treated as negatives. The instance score is set to 1 for positives and 0 for negatives, which is used to compute instance score regression loss.


In inference, the detection module detects all classes of object instances firstly. Then, each detected bbox is forward to the attention model for generating instance-specific features covering a particular instance. Finally, all instance masks are segmented separately.

\section{Experiments}

\begin{figure*}
\centering
\includegraphics[height=9cm]{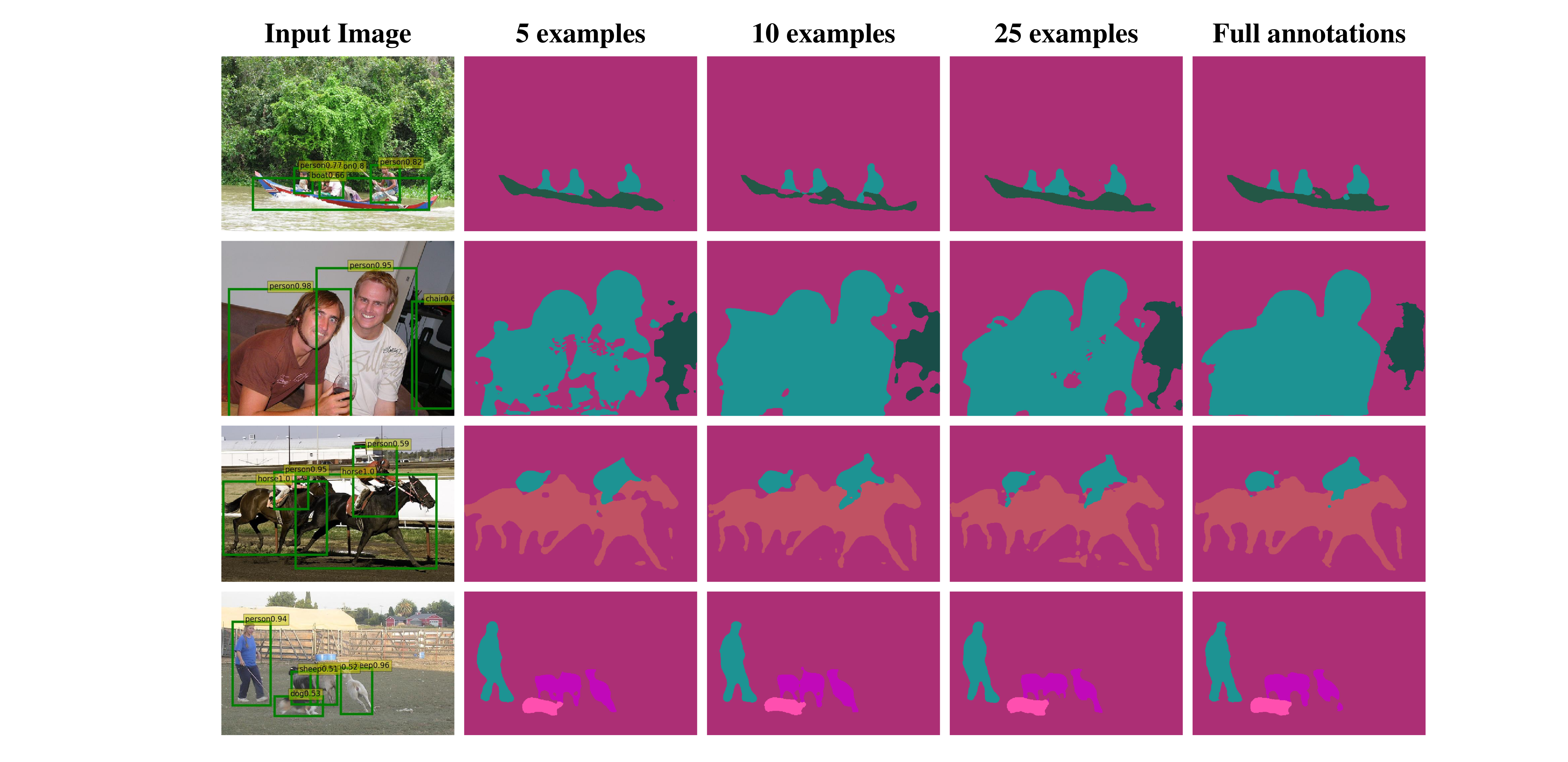}
\caption{Semantic segmentation results of several examples from DASNet on VOC test dataset. The DASNet is trained with 16.5K bbox-level annotated images, and different number of pixel-level annotations, including 5, 10, 25 examples per class, and 10K (full) examples in total.}
\label{fig:seg}
\end{figure*}

\begin{figure*}
\centering
\includegraphics[height=5cm]{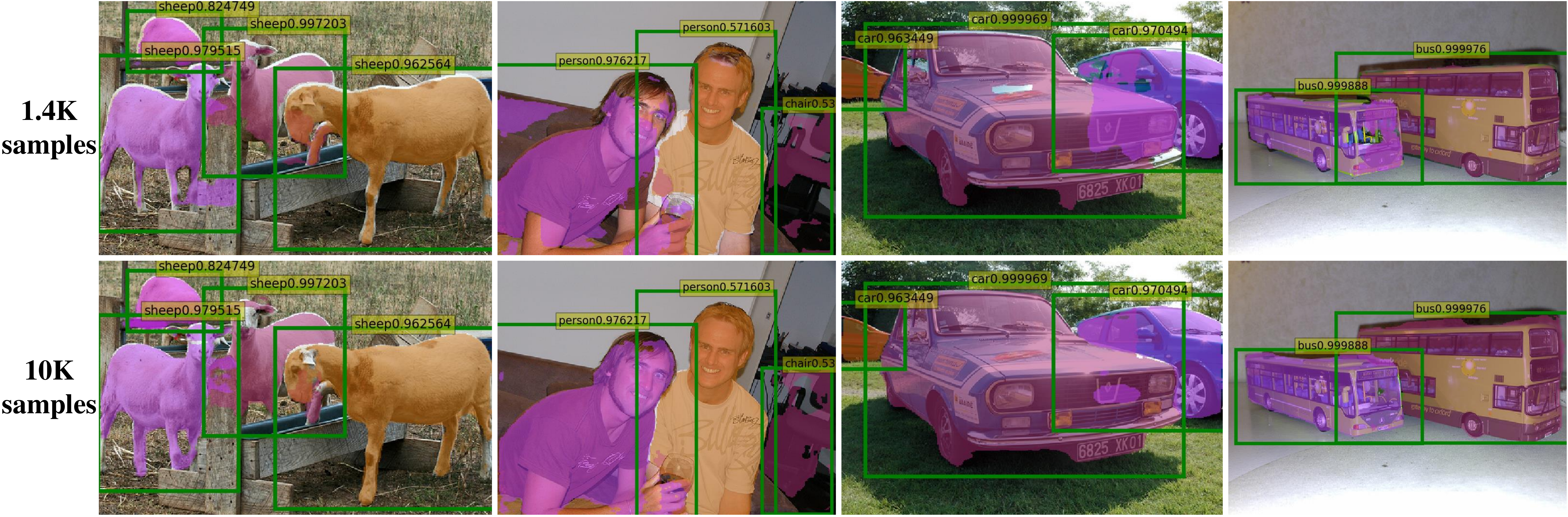}
\caption{Instance segmentation results of several examples from DASNet on VOC test dataset. The DASNet is trained with 145K bbox-level annotated images, and different number of pixel-level annotations, including 1.4K and 10K examples in total.}
\label{fig:inst}
\end{figure*}

\subsection{Implementations Details.}

\textbf{Dataset:}
In our experiments, we focus on the 20 Pascal classes \cite{Everingham2010The}. We employ PASCAL VOC and MS COCO \cite{Lin2014Microsoft} datasets for training and PASCAL VOC 2012 dataset for testing. The PASCAL dataset is extented with 10528 pixel-level annotated images in \cite{Hariharan2011}. To simulate semi-supervised learning scenario, we construct a heterogeneous annotated training set, in which all images are labeled with bbox annotations and a fraction of images have pixel-level annotations. A various number of bbox and mask annotations of training samples is controlled to demonstrate the effectiveness of our semi-supervised framework. To compare with existing weakly-, semi-, and fully-supervised learning methods \cite{Papandreou2015}, \cite{Hong2015Decoupled}, \cite{Dai2015}, \cite{Khoreva2017}, \cite{Chen2014Semantic}, \cite{Long2015Fully}, we split mask annotations to $Seg10K$ (10528 images in \cite{Hariharan2011}), $Seg1.4k$ (1464 images in VOC 2012 training dataset), $Seg500, Seg200, Seg100$ (25, 10, 5 images per class random select from $Seg10K$), and split the bbox annotations into $Box16.5K$ (16551 images from VOC2007 trainval and VOC2012 trainval datasets), $Box145K$ (144790 images from $Box16.5K$, VOC2007 test and MS COCO datasets). All images are resized to $320 \times 320$ during training and testing\footnote{On $Box145K$, the images are resized to $300 \times 300$ for training the detection module since we use the released SSD model from \emph{https://github.com/weiliu89/caffe/tree/ssd}.}.

\textbf{Data Augmentation:}
We use common strategies to augment training samples, including expending, cropping, adding noise, and horizontal flipping as in \cite{Liu2016SSD}. In addition, the number of class-specific bboxes is randomly set, which is similar to the combinatorial cropping proposed in \cite{Hong2015Decoupled}.

\textbf{Optimization:}
Our proposed method is implemented based on Caffe library \cite{jia2014caffe}. To compare with existing methods, we use VGG 16-layer net \cite{Simonyan2014Very} as our backbone. We use the standard Stochastic Gradient Descent (SGD) with momentum for optimization, where the base learning rate is 0.01 for semantic segmentation and 0.02 for instance segmentation, and divided by 10 after 15K/30K, the batch size is 32, and the momentum is 0.9. The source code will be available.

\begin{table}
\renewcommand{\arraystretch}{1.2}
\renewcommand\tabcolsep{3.5pt}
\begin{center}
\caption{Semantic segmentation results (IoU scores) on VOC 2012 test dataset.}
\label{table:semantic}
\begin{tabular}{c|c|ccc|c}
Sup. & Method & \#Mask & \#BBox & \#Tag & Test \\
\hline
\noalign{\smallskip}
\hline
\multirow{5}{*}{Weak}  & WSSL$_R$ \cite{Papandreou2015} & - & \multirow{5}{*}{10K}   & - & 54.2 \\
                       & WSSL$_S$ \cite{Papandreou2015} & - &                        & - & 62.2 \\
                       & BoxSup$_{MCG}$ \cite{Dai2015}  & - &                        & - & 64.6 \\
                       & Box$^i$ \cite{Khoreva2017}     & - &                        & - & 67.5 \\
                       & M $\cap$ G+ \cite{Khoreva2017} & - &                        & - & 67.5 \\
\hline
\multirow{2}{*}{Fully} & FCN \cite{Long2015Fully}              & \multirow{2}{*}{10K} & - & - & 62.2  \\
                        & DeepLabCRF \cite{Chen2014Semantic}    &           & - & - & 66.4 \\ \cline{2-6}

\hline
\multirow{11}{*}{Semi} & \multirow{4}{*}{DecoupledNet \cite{Hong2015Decoupled}} & 100 & - & \multirow{4}{*}{10K}  & 54.7  \\
                      &                                                        & 200 & - &  & 58.7 \\
                      &                                                        & 500 & - &  & 62.5 \\
                      &                                                        & 10K & - &  & 66.6 \\ \cline{2-6}
                      & BoxSup$_{MCG}$ \cite{Dai2015} & \multirow{3}{*}{1.4K} & \multirow{3}{*}{9K} & - & 66.3 \\
                      & WSSL$_s$ \cite{Papandreou2015} &  &  & - & 55.5 \\
                      & M $\cap$ G+ \cite{Khoreva2017} &  &  & - & 66.9 \\  \cline{2-6}
                      & \multirow{4}{*}{\textbf{DASNet}}              & 100 & 16.5k& - & \textbf{64.8} \\
                      &                         & 200 & 16.5k& - & \textbf{66.6} \\
                      &                         & 500 & 16.5k& - & \textbf{67.9} \\
                      &                         & 10K & 16.5k& - & \textbf{69.2} \\
\hline
\end{tabular}
\end{center}
\end{table}

\subsection{Semantic Segmentation}
\label{sec:semantic}
In this section, we evaluate the performance of DASNet for semantic segmentation task on VOC 2012 test set via evaluation server. Segmentation accuracy is measured by Intersection over Union (IoU) between ground-truth and predicted segmentation. Table \ref{table:semantic} compares quantitative results of using various supervision level.

Training with a small number of pixel-level mask annotations, the DASNet presents substantially better performance without any post-processing than other weakly-, semi-, and fully-supervised methods. Particularly, when five training examples per class with mask annotations are used, the accuracy of semi-supervised method DecoupledNet is reduced by 11.9\% (66.6-54.7\%), while the accuracy of our DASNet is reduced by 4.4\% (69.2-64.8\%) only, comparing with using the full mask annotations. The results show that the DASNet can significantly reduce the pixel-level training data requirements for semantic segmentation, although we use stronger bbox-level annotations than the image-level. Compared with semi-supervised methods which use the same type of training examples, the DASNet also requires much less pixel-level annotations for achieving similar results. Moreover, the DASNet trained with 200 strong annotations and 16.5K bbox annotations can obtain higher accuracy than some fully supervised methods trained with 10K strong annotations.

Fig. \ref{fig:seg} presents some qualitative semantic segmentation results produced by our DASNet on VOC 2012 test set. Trained with an extremely small number (5-10 examples per class) of pixel-level annotated samples only, the DASNet can also segment multiple objects of various sizes.


\begin{table}
\renewcommand{\arraystretch}{1.2}
\renewcommand\tabcolsep{2.8pt}
\begin{center}
\caption{Instance segmentation results on VOC 2012 validation set.}
\label{table:instance}
\begin{tabular}{c|c|cc|cc}
Sup. & Method & \#Mask & \#BBox & mAP$^r_{0.5}$ & mAP$^r_{0.7}$ \\
\hline
\noalign{\smallskip}
\hline
\multirow{4}{*}{Weak}  & \multirow{2}{*}{DeepMask \cite{Khoreva2017}}              & - & 10K  & 39.4 & 8.1  \\
                       &                                                           & - & 110K & 42.9 & 11,5\\
                       & \multirow{2}{*}{DeepLab$_{BOX}$ \cite{Khoreva2017}}       & - & 10K  & 44.8 & 16.3 \\
                       &                                                           & - & 110K & 46.4 & 18.5\\
\hline
\multirow{4}{*}{Fully} & \multirow{2}{*}{DeepMask \cite{Khoreva2017}}              & 10K  & - & 41.7 & 9.7 \\
                       &                                                           & 110K & - & 44.7 & 13.1 \\
                       & \multirow{2}{*}{DeepLab$_{BOX}$ \cite{Khoreva2017}}       & 10K  & - & 27.5 & 20.2 \\
                       &                                                           & 110K & - & 49.4 &  23.7\\
\hline
\multirow{2}{*}{Semi}  & \multirow{2}{*}{DASNet}               & 1.4K & 145K  & \textbf{56.2} &\textbf{30.5} \\
                       &                                       & 10K  & 145K  & \textbf{57.6} & \textbf{33.7} \\
\hline
\end{tabular}
\end{center}
\end{table}


\subsection{Instance segmentation}
Following \cite{Khoreva2017}, we evaluate the performance of DASNet for instance segmentation task with mAP$^r$ at IoU threshold 0.5 and 0.75. Table \ref{table:instance} shows the instance segmentation results. As the number of pixel training samples is reduced from 10K to 1.4K, the instance segmentation accuracy is reduced by 1.4\% only, which shares the similar results with semantic segmentation task. Fig. \ref{fig:inst} shows several examples of instance segmentation from VOC 2012 test dataset .

\section{Discussion}

\begin{figure}
\centering
\includegraphics[height=4.5cm]{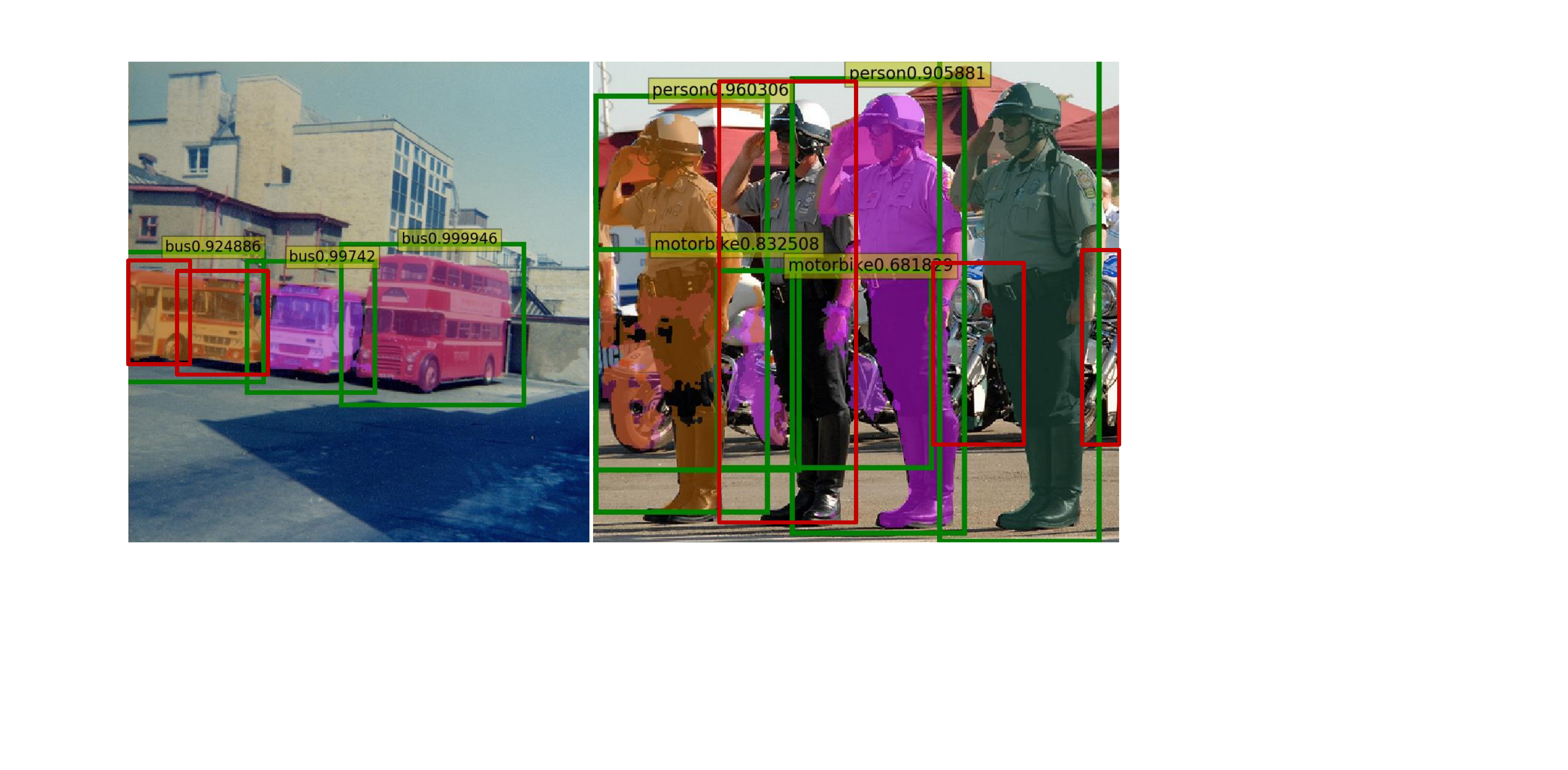}
\caption{Failed examples in instance segmentation. The red bboxes are not detected correctly by the detection module.}
\label{fig:failed}
\end{figure}
By exploring the detection model, our DASNet significantly improves the performance of semi-supervised instance and semantic segmentation compared with existing methods. The proposed detection-attention-segmentation framework is actually a from-coarse-to-fine localization process. Thus the large scale available bbox annotations can improve segmentation results in an accurate and efficient way by training the detection model. However, this also raises the fore-end error problem that the detection errors will directly lead to segmentation errors, as shown in Fig. \ref{fig:failed}. In future works, we have three suggestions for this problem. First, integrating the state-of-the-art detection model into the DASNet framework to reduce the detection error. Second, obtaining the class-specific feature maps with pixel-level resolution instead the bbox-level in this work. Third, joint learning is explored to further improve the performance of both detection and segmentation modules. These suggestions will be included in our future research and more experiments should be conducted for further demonstrating the effectiveness of the proposed DASNet.

\section{Conclusions}
This paper introduces DASNet, a semi-supervised instance and semantic segmentation framework for reducing pixel-level mask annotations by leveraging large scale bbox annotations. The key idea is exploring detection models to simplify the pixel-level supervised learning task. Thus the pixel-level training data requirements of segmentation model are reduced. The attention module is a key component that exploits the products of detection to facilitate learning class-agnostic segmentation by multi-scale class-specific feature maps. In addition, the position-sensitive score map technique is adapted to DASNet for instance segmentation. Experimental results show that our method substantially reduces the requirements of pixel-level annotations compared with existing semi-supervised instance and semantic segmentation methods.

\section{Acknowledgements}
The research was supported by the National Natural Science Foundation of China (61571353) and the Science and Technology Projects of Xi’an, China (201809170CX11JC12).

{\small
\bibliographystyle{ieee}
\bibliography{DASNet}
}

\end{document}